\documentclass{midl} 

\usepackage{mwe} 
\jmlrvolume{-- Under Review}

\usepackage{url}            
\usepackage{booktabs}       
\usepackage{amsfonts}       
\usepackage{nicefrac}       
\usepackage{microtype}      
\usepackage{enumitem}
\PassOptionsToPackage{dvipsnames}{xcolor}
\usepackage[dvipsnames]{xcolor}
\usepackage{bm}
\usepackage{amsmath}
\usepackage{multirow}
\usepackage{threeparttable}
\usepackage{adjustbox}
\usepackage{eqparbox}
\usepackage{arydshln}
\usepackage{wrapfig}

\usepackage{caption}
\usepackage{graphicx}
\usepackage{amsmath,amssymb} 
\usepackage{multirow}

\usepackage{hyperref}
\hypersetup{
    colorlinks,
    linkcolor={red!50!black},
    citecolor={blue!50!black},
    urlcolor={blue!80!black}
}

\title[Using LTI Models with Structured Prompts for Skin Disease Identification]{Using Large Text-to-Image Models with Structured\\ Prompts for Skin Disease Identification: A Case Study}

\midlauthor{\Name{Sajith Rajapaksa\nametag{$^{1,3}$}} \Email{sajith@yukoai.com}
 \AND
\Name{Jean Marie Uwabeza Vianney\nametag{$^{1}$}} \Email{jean@yukoai.com}
 \AND
\Name{Renell Castro\nametag{$^{1}$}} \Email{renell@yukoai.com}
 \AND
\Name{Farzad Khalvati\nametag{$^{2,3,4}$}} \Email{farzad.khalvati@utoronto.ca}
 \AND
\Name{Shubhra Aich\nametag{$^{1,5}$}} \Email{saich@andrew.cmu.edu}\\
\addr $^{1}$ Yuko AI Inc. Toronto, ON, Canada \\
\addr $^{2}$ University of Toronto, ON, Canada\\
\addr $^{3}$ SickKids Hospital, Toronto, ON, Canada\\
\addr $^{4}$ Vector Institute,  Toronto, ON, Canada\\
\addr $^{5}$ Robotics Institute, Carnegie Mellon University, Pittsburgh, PA, USA
}

\begin{document}

\maketitle

\vspace{-2ex}
\begin{abstract}
This paper investigates the potential usage of large text-to-image (LTI) models for the automated diagnosis of a few skin conditions with rarity or a serious lack of annotated datasets. As the input to the LTI model, we provide the targeted instantiation of a generic but succinct prompt structure designed upon careful observations of the conditional narratives from the standard medical textbooks. In this regard, we pave the path to utilizing accessible textbook descriptions for automated diagnosis of conditions with data scarcity through the lens of LTI models. Experiments show the efficacy of the proposed framework including much better localization of the infected regions. Moreover, it has the immense possibility for generalization across the medical sub-domains, not only to mitigate the data scarcity issue but also to debias automated diagnostics from the all-pervasive racial biases. 
\end{abstract}

\begin{keywords}
Large text-to-image model, structured prompts, textbook descriptions, skin disease detection.
\end{keywords}

\section{Introduction}
\label{sec:intro}

The limited availability of large annotated datasets is a serious impediment to progress in AI for medical imaging. In recent years, large pretrained text-to-image (LTI) generative models, such as  DALL-E2 \cite{ramesh2022hierarchical}, Imagen \cite{saharia2022photorealistic} have been used to greatly expedite the development of AI applications by eradicating the bottleneck of large-scale annotated datasets \cite{kaliamoorthi2021distilling}. These models are trained on a massive amount of data using contextual unsupervised or self-supervised learning techniques. Such contextual training helps the models to learn the patterns and structures present in large amounts of text and image data and capture the contextual relationships between words and their meanings. \vspace{-1.5ex}
\\~\\ Although these LTI models are exposed to diverse datasets while training, the extent of the nature of data used for training is unknown. Moreover, how much diversity is captured in the pretrained weights is also an open question. Consequently, it is not clear whether these models encountered or assimilated the information related to skin conditions sufficient enough to be useful for automated diagnosis. 
Recent work shows the potential of realistic image generation to fool the classifier trained on real images \cite{stanford-xray-diffusion}. However, in case of little or no real samples, the possibility of developing medical AI applications only with these LTI generated samples is still an uncharted territory. \vspace{-1.5ex}
\\~\\ Therefore, in this paper, we provide \textbf{an initial case study regarding the usage of LTI models for a few skin conditions for which the annotated dataset required for AI-assisted automation is scarce}. In particular, we employ DALL-E mini \cite{Dayma_DALL_E_Mini_2021} to generate representative images for the four skin conditions -- Atopic dermatitis, Urticaria Hives, Scabies, and Warts. The inputs to models like DALL-E and its variants are the text prompts describing the properties of the image to be generated. Providing effective inputs to these models is an art on its own, also known as prompt engineering \cite{liu2022design}. Recently, prompt engineering is receiving a lot of attention from the research community to devise better prompts for more targeted image generation and refinement \cite{liu2022design,zhou2022learning}. However, in this paper, for our very initial case study, we stick to the simple (manual) formats equipped with disease names, skin tones, and symptoms and characteristics taken from the standard medical textbooks to assess the feasibility of these large models for the diagnosis of conditions. In this regard, we also investigate \textbf{the research question of utilizing textbook narrations, which are easy to obtain, for automated diagnosis of comparatively rare conditions through the lens of LTI models.} \vspace{-1.5ex}
\\~\\
We show that the models trained with LTI images improve the classification accuracy of skin conditions over the model trained only on a scarce real dataset. Moreover, further visualization of the activation maps on real data indicates much better localization of the infected areas for the model trained with LTI-generated images compared to baseline (Figure \ref{fig:cam} and supplementary material). We believe such a substantial improvement in terms of localization would help in accelerating medical imaging research for which segmentation and localization play a big role in general. 
To the best of our knowledge, this is the first work regarding a strategic framework to utilize LTI models equipped with textbook based structured prompts for improved medical image analysis where the need for training data is an issue that is commonplace in this domain. \vspace{1ex}

\noindent Overall, our contributions are as follows: \vspace{-1.5ex}
\begin{figure}[]
    \centering
    \includegraphics[width=\linewidth]{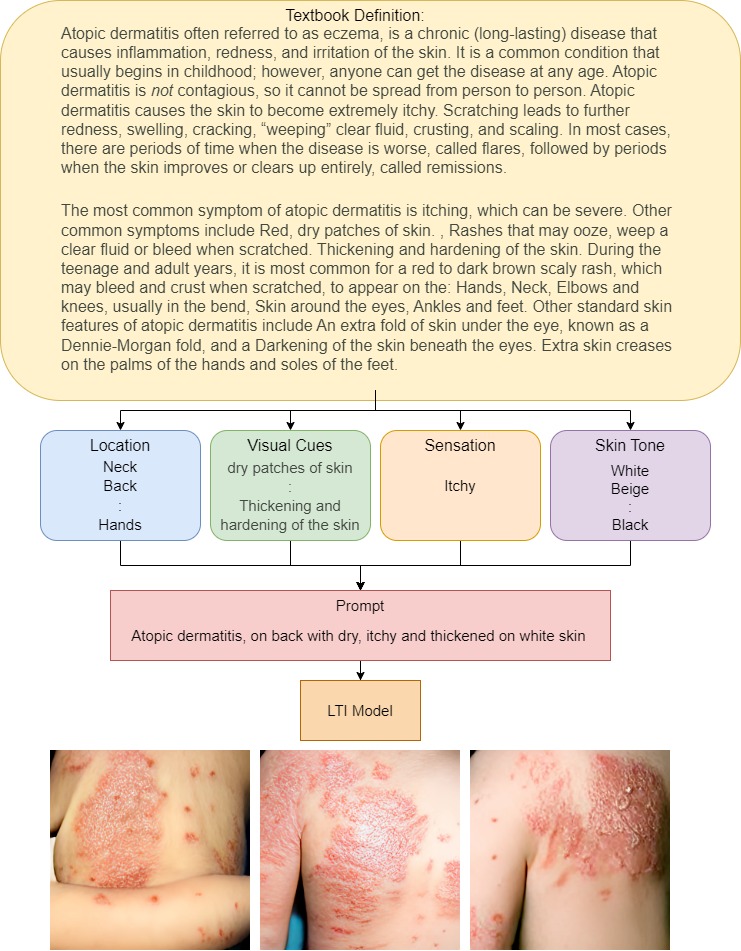}
    \caption{The proposed framework for generating images tailored to particular skin conditions based on textbook descriptions. First, the long textbook narrative is parsed into a few keywords following our generic but succinct prompt structure. Next, this prompt instantiation is fed into the LTI model to generate targeted images for further training.}
    \label{fig:gen}
\end{figure}

\begin{itemize}[leftmargin=*]
    \item Our initial case study is the first known attempt to analyze the impact of LTI models for (comparatively rare) skin conditions through the lens of a strategic framework with a tentative guideline for simple prompt engineering based on medical textbooks.\vspace{-1.5ex} 
    \item We show that the deep learning model trained on disease classification only with LTI generated images exhibits much better localization of the disease ROI in terms of class activation maps, thus demonstrating the efficacy of the proposed framework. \vspace{-1.5ex}
    \item Although the study in this paper is based on skin conditions, our framework is domain-agnostic to be applied to any sub-domain in medical image analysis dealing with the scarcity of targeted training data.\vspace{-1.5ex} 
    \item Our framework with tentative prompt engineering guidelines can be easily extended by the community to significantly resolve the issue of racial bias (and potentially other prevalent ones) in medical datasets.
\end{itemize}

\section{Methods}

Our framework is depicted in Figure 1. It comprises two main components -- (1) The choice of a set of structured text prompts used as inputs for data generation, and (2) training a classification model using the synthetic images generated by the prompts. 

\subsection{Image Generation}

Our objective is to utilize the textbook descriptions of the skin conditions to generate images for training the AI models. To do so, we employ an LTI model that generates images based on the text prompts provided as inputs. However, the textbook descriptions are usually quite long and attempt to describe the whole condition in a generalized manner. Although such descriptions are good for human-level communication, we find it somewhat difficult to generate targeted images simply using these long narratives. \vspace{-1.5ex}
\\~\\
Consequently, a succinct structure of the text prompts comprising placeholders for only the target set of keywords is required. The predefined prompt structure makes it generalizable across medical conditions at least in a particular sub-domain like skin conditions. The placeholders within the generic structure can be replaced with the words sufficient and necessary to properly identify each condition. \vspace{-1.5ex}
\\~\\
To develop such a generic but laconic structure, we start by analyzing the textbook definitions of skin conditions. These definitions serve as the basis for the image generation process and help us understand the key features and characteristics of the conditions. We find that almost all these definitions contain four global features: \vspace{-1.5ex}

\begin{itemize}[leftmargin=*]
	\item Physical location of the occurrence of disease within the body. \vspace{-1.2ex}
	\item Skin tone of the affected individual following the Fitzpatrick scale \cite{roberts2009skin}.\vspace{-1.2ex}
	\item Visual cues, such as color, shape, appearance, etc.\vspace{-1.2ex}
	\item Sensation in or around the affected region. Note that sometimes the sensation is responsible for the partial change in appearance depending upon the actions taken by the patients to deal with that. Thus, despite being a completely physical feeling, including this information helps in better image generation.
\end{itemize}

\vspace{-1ex} \noindent Following this observation, our generic prompt structure takes the form of \vspace{-1ex}

\begin{center} 
\textcolor{teal}{\texttt{<VISUAL\_CUES> + <SENSATION> + <PHYSICAL\_LOCATION> + <SKIN\_TONE>}}
\end{center}

\vspace{-1ex}\noindent An instantiation of our prompt based on the varying \texttt{<SKIN\_TONE>} is shown in Figure 2. \vspace{-1.5ex}
\\~\\
\noindent Finally, to generate images with this structure, we instantiate the placeholder tags with concrete descriptions and feed the instantiations into an LTI model with different random seeds. We employ DALL-E mini \cite{Dayma_DALL_E_Mini_2021} as our LTI candidate since it was fully open-sourced at the time of writing this paper.

\begin{figure}
    \centering
    \includegraphics[width=16cm]{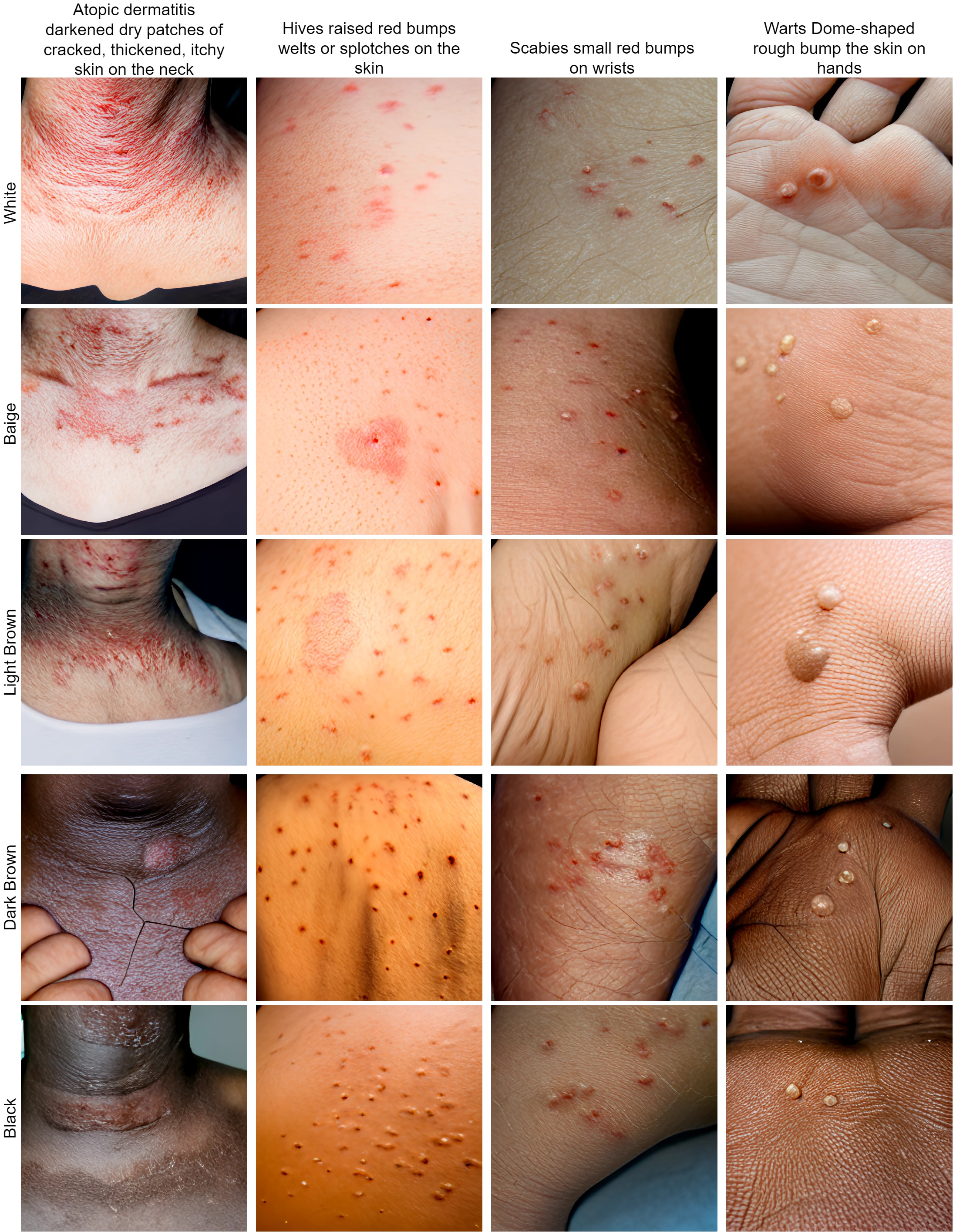}
    \caption{Samples generated by varying the \texttt{<SKIN\_TONE>} parameter (left) of our prompt. The rest of the prompt instantiation is on the top of each column.}
    \label{fig:gen}
\end{figure}

\subsection{Classification Model}

Using the images generated with the set of instantiated prompts, we train a standard image classification model (ResNet50 \cite{he2016deep} pretrained on ImageNet) for disease classification. This classifier is then evaluated on the real dataset both quantitatively and qualitatively. Details are provided in the Experiments section below. 

\begin{figure}[h]
    \centering
    \includegraphics[width=13cm]{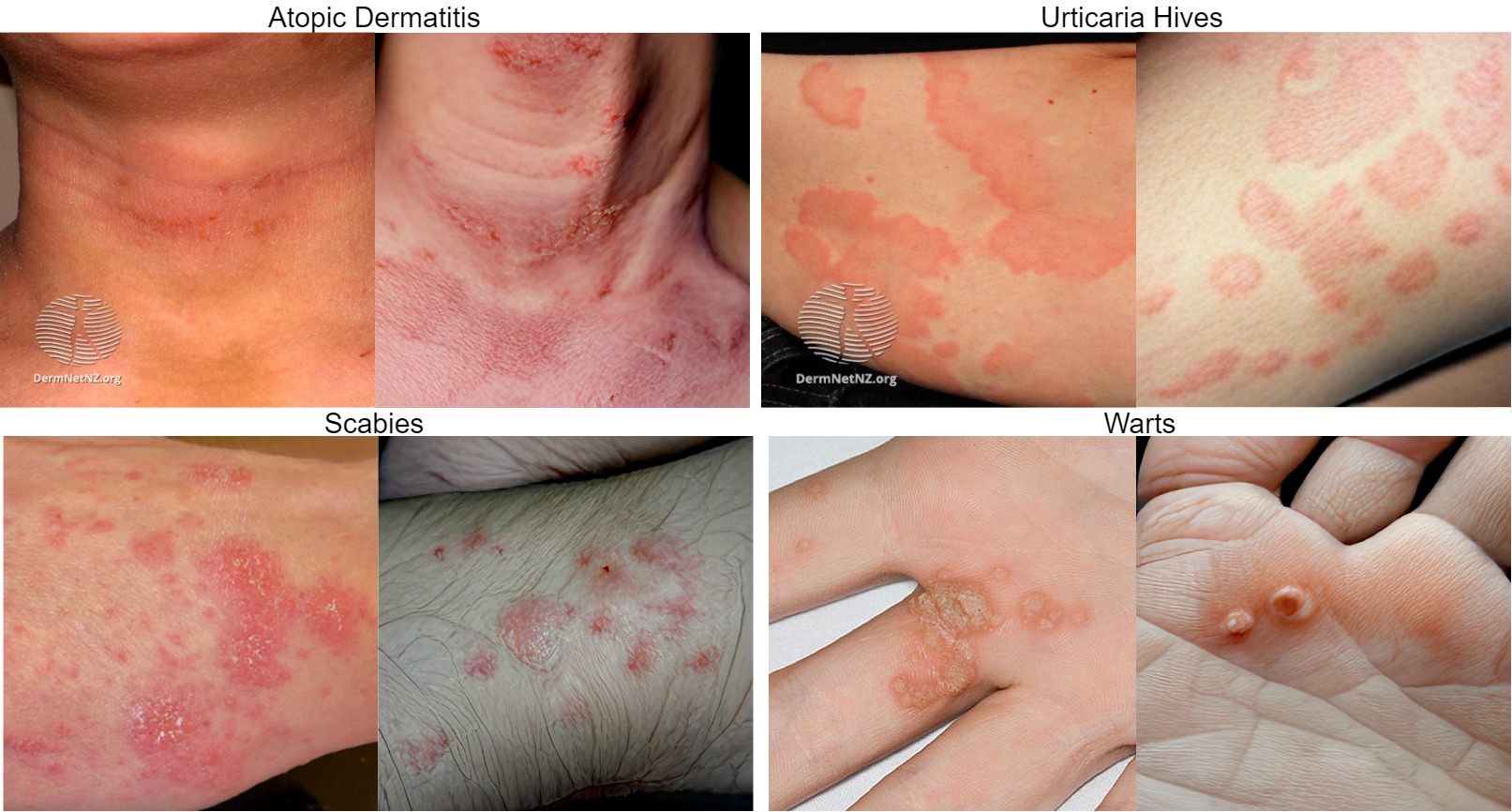} \vspace{-1ex}
    \caption{Sample comparison between real \cite{Dermnet} (left) and generated images (right) for each of the four skin conditions.\vspace{-1ex}}
    \label{fig:gen}
\end{figure}

\section{Experiments}

\noindent\textbf{Dataset:} We generate images for skin conditions with comparatively rare data availability but for which there are at least some real samples available for evaluation. Such availability of a small amount of real data helps to quantitatively evaluate the impact of the generated images for diagnosis. Thus, four skin conditions are selected for our experiments -- (1) Atopic dermatitis, (2) Scabies, (3) Urticaria hives, and (4) Warts. For each of these four categories, $\sim 1K$ images are synthesized with varying physical locations and skin tones for training. A comparison between the real and synthetic images is shown in Figure \ref{fig:gen}. Additional synthetic examples are provided in the supplementary material. \vspace{-2.0ex}
\\~\\
The real dataset used to test our framework is the publicly available Dermnet \cite{Dermnet} containing 115, 55, 66, 131 samples for Atopic dermatitis, Scabies, Urticaria hives, and Warts, respectively. These samples are randomly split into 10\% for finetuning and 90\% for evaluation. Note that the subset for finetuning is used to show the potential improvement over the synthetic training with a small amount of real data. \vspace{-2.0ex}
\\~\\
We could not find any real samples with \texttt{non-white} skin tones for evaluation in this paper whereas our synthetic training set is balanced in this regard, thanks to our generic prompt structure. Thus, our evaluation (not training) has an unavoidable racial bias all-pervasive in the medical domain that we wish to eradicate in future.\vspace{-1.0ex}
\\~\\
\textbf{Training and Implementation: } We train the ResNet50 \cite{he2016deep} architecture for classification in PyTorch for 50 epochs with a learning rate of $1e^{-4}$ using Adam optimizer \cite{kingma2014adam} and cross-entropy loss. The logos in Dermnet images were removed with standard image preprocessing techniques \cite{yu2018generative, yu2018free}. This is to ensure consistency and eradicate any potential confounding factors that could affect the performance of the classifier.  \vspace{-1.0ex}
\\~\\
\textbf{Evaluation:} All the evaluations in this paper are performed on the real dataset. For quantitative evaluation, first, we assess the model's classification accuracy trained only on our synthetic data to get a sense of the synthetic to real generalization ability. Next, we re-evaluate the synthetically trained model after finetuning it on a few real samples -- $\sim 10$ from each category. This is to see the potential improvement that a tiny real dataset brings to the table alongside the synthetic dataset. Moreover, we also visualize the class activation maps \cite{selvaraju2017grad} to analyze the saliency of the learned model. 

\begin{table}[h]
\centering
\caption{Comparison between synthetic and real fine-tuning. \vspace{-1.5ex}}
\label{tab:comp}
\begin{threeparttable}
\begin{tabular}{l|ccc} 
\hline
 & ImageNet + LTI & ImageNet + Finetune & ImageNet + LTI + Finetune \\ \cline{2-4} 
Accuracy ($\%$) & 42.0 & 56.0 $\pm{0.6}$ & 63.0 $\pm{2.8}$ \\ 
\hline
\end{tabular}
\begin{tablenotes}
\item[] \small \hspace{-2ex} The deviations $(\pm{})$ are reported over the average of 10 runs. \vspace{-2ex}
\end{tablenotes}  
\end{threeparttable}
\end{table}

\begin{table}[h]
\centering
\caption{Normalized confusion matrix for the synthetic only model. \vspace{-1.5ex}}
\label{tab:nconf}
\begin{tabular}{l|cccc}
\hline
\multicolumn{1}{c|}{} & \multicolumn{4}{c}{Prediction} \\ \cline{2-5} 
\multicolumn{1}{c|}{\multirow{-2}{*}{Ground truth}} & Atopic dermatitis & Urticaria hives &  Scabies & Warts \\ \hline
Atopic dermatitis & \textbf{0.64} & 0.05 & 0.27 & 0.04 \\
Urticaria hives & 0.37 & \textbf{0.47} & 0.14 & 0.02 \\
Scabies & 0.36 & 0.04 & \textbf{0.48} & 0.12 \\
Warts & \textcolor{red}{\textbf{0.53}} & 0.01 & 0.32 & 0.14 \\ \hline
\end{tabular}
\end{table}

\begin{figure}[h]
    \centering
    \includegraphics[width=11.8cm]{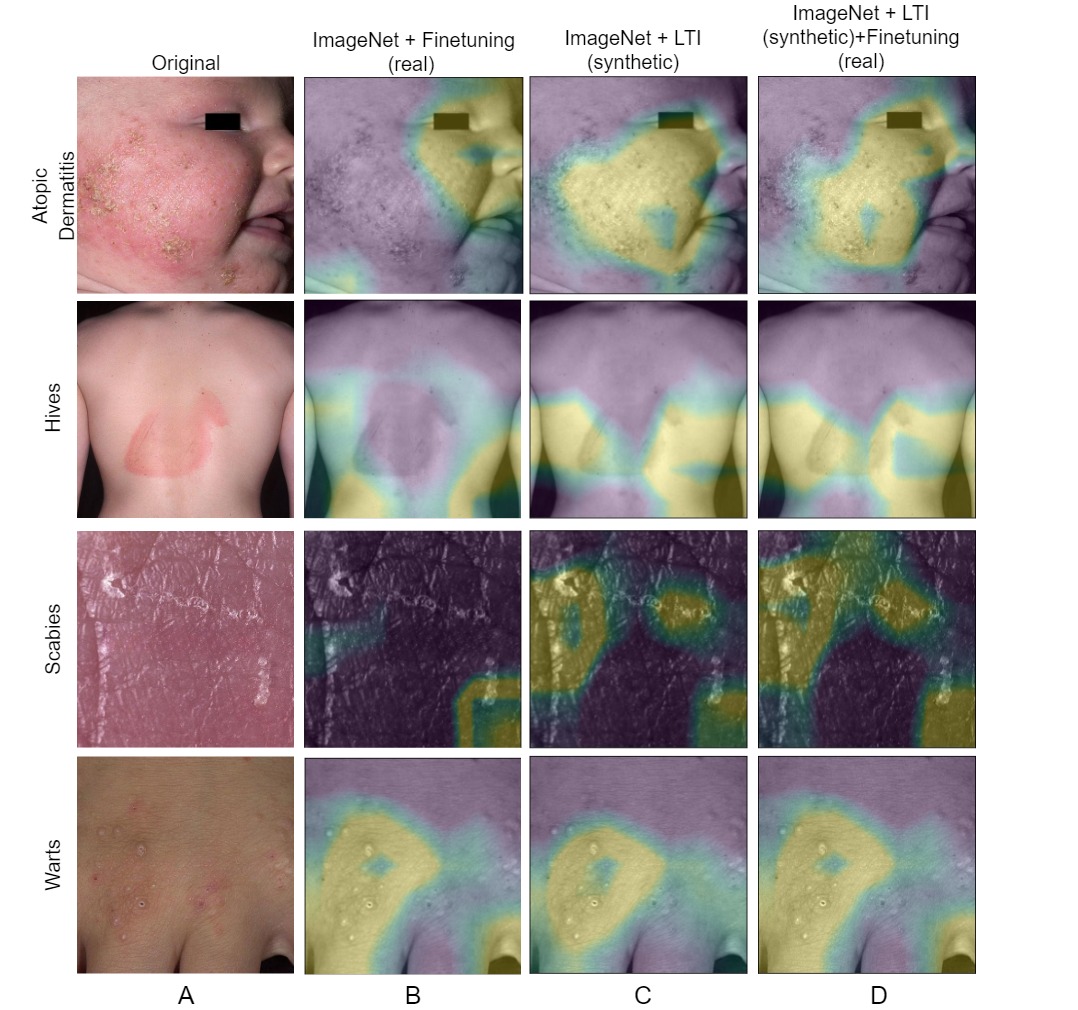}
    \vspace{-1ex}
    \caption{Comparison of the class activation maps for different protocols -- (A) Images; (B) ImageNet + Finetuning (real); (C) ImageNet + LTI (synthetic); (D) ImageNet + LTI (synthetic) + Finetuning (real). The activation with the synthetic training (C and D) is qualitatively more accurate than the one without it (B).\vspace{-0.2cm}}
    \label{fig:cam}
\end{figure}

\vspace{-0.0ex}
\noindent \textbf{Results:} The LTI generated images (Figure \ref{fig:gen}) capture the visual characteristics distinctive to individual skin conditions more or less well. Training the disease classification model only with these generated images achieves 42\% accuracy on the 4 class classifications (Table \ref{tab:nconf}). From the normalized confusion matrix of this synthetic-only results shown in Table \ref{tab:nconf}, it is clear that the comparatively lower accuracy is mostly attributed to the \texttt{Warts} class being misclassified into \texttt{Atopic dermatitis}. This is because of the significant visual similarity of these two classes in the real dataset \cite{Dermnet} used for evaluation. Despite the modest classification accuracy, the quality of disease localization shows significant improvement compared to baseline finetuning (see Figure \ref{fig:cam} and supplementary material). \vspace{-2.0ex}
\\~\\
Moreover, we finetune the logit layer of the base model (ImageNet pretrained) and synthetic one on a small random subset of the real samples (~10 per class). The results over 5 random runs are reported in Table \ref{tab:comp}. The synthetically trained model improves the baseline model accuracy by 7.0\% (63.0 \textit{vs.} 56.0) on average. \vspace{-2.0ex}
\\~\\
Also, sample comparisons of the class activation maps \cite{selvaraju2017grad} are shown in Figure \ref{fig:cam} among the synthetic model, the baseline, and the synthetic with real finetuning. In this figure, the quality of saliency detection after the synthetic training looks much better than the one without this step. Thus, both from the numerical and qualitative comparisons, the efficacy of the synthetic examples is evident. We hypothesize that the accuracy of the synthetic-only model can be improved further by combining our structured prompt generation with non-redundant sampling strategies. Please consult the supplementary material for additional visualizations. \vspace{-1ex}

\section{Conclusion and Future Work}
\label{sec:conclusion}

In this paper, as an initial case study, we demonstrate the potential of the LTI models to be promising for skin disease detection for which there is a serious lack of training data. At this point, we only employ the images generated with structured text prompts equipped with textbook descriptions for training, and disease names or tags just for labeling. In this regard, our first attempt is somewhat unimodal using only the generated images. However, the text prompts enriched with the laconic description following the textbooks provide an additional avenue to explore multimodal learning for improved performance and explainability. As part of future work, we will extend the methods to more diseases alongside the multimodal learning mentioned above. Moreover, the parsing of the textbook narrative into the prompt is performed manually in this paper. Automation of this process guided by the generic prompt structure will streamline application development. In addition, the recent advances in finetuning the diffusion models \cite{stanford-xray-diffusion} for better visual data generation can also be explored as a future direction. \vspace{-2.0ex}
\\~\\
Most importantly, skin conditions appear differently based on skin tones and race. This issue is arguably among the most difficult ones to resolve with real datasets in the foreseeable future. The explicit and lucid nature of the text prompts (e.g. one particular tag for skin tone) used for data generation in this paper seems to be a promising way to mitigate such all-pervasive racial biases in the AI-assisted medical imaging domain. At the same time, such racial debiasing would also help us to create hyperlocal automated assistance programs for underserved communities in remote areas in alignment with our core organizational principles.\vspace{-2.0ex}
\\~\\
Finally, regarding generalization, although we are focusing on skin diseases for organizational purposes, we believe this study with a tentative guideline for prompt engineering will encourage the research community to utilize similar frameworks for other medical conditions in general as well.


\bibliography{midl}
\newpage
\section*{Supplementary Material}
\begin{figure}[h]
    \centering
    \includegraphics[width=14cm]{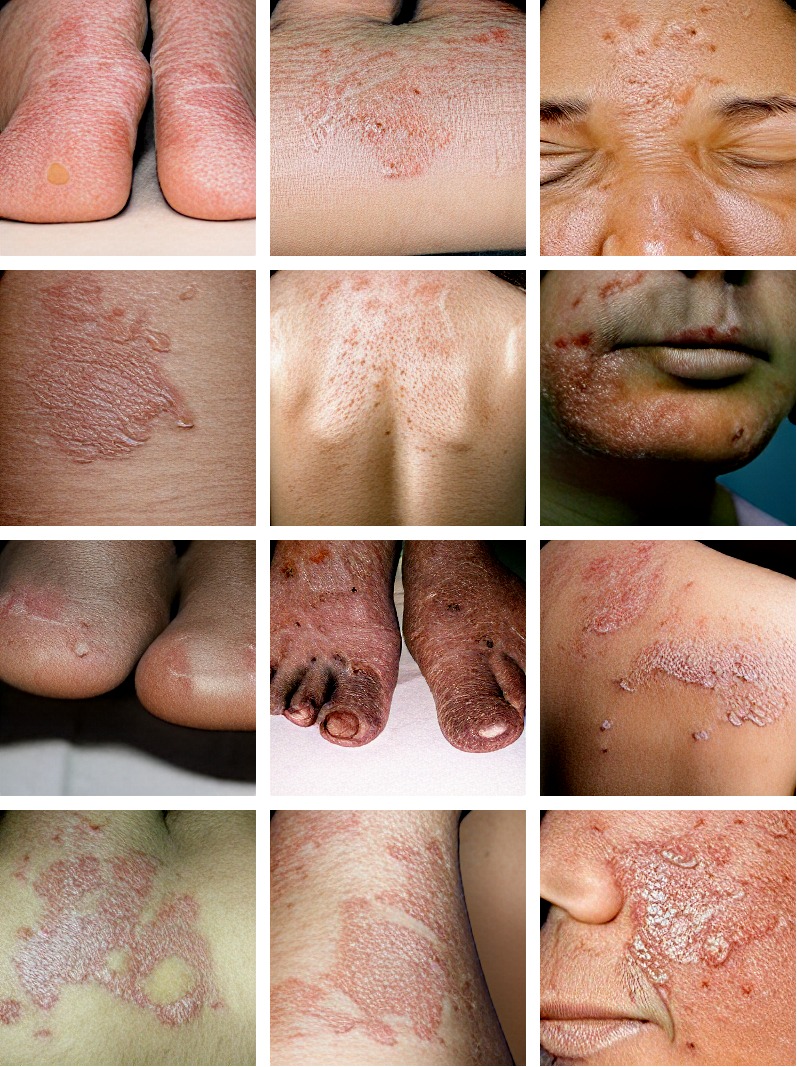}
    \caption{LTI generated samples for \texttt{Atopic dermatitis} with our structured prompts. }
    \label{fig:gen_supp_1}
\end{figure}
\begin{figure}[t]
    \centering
    \includegraphics[width=14cm]{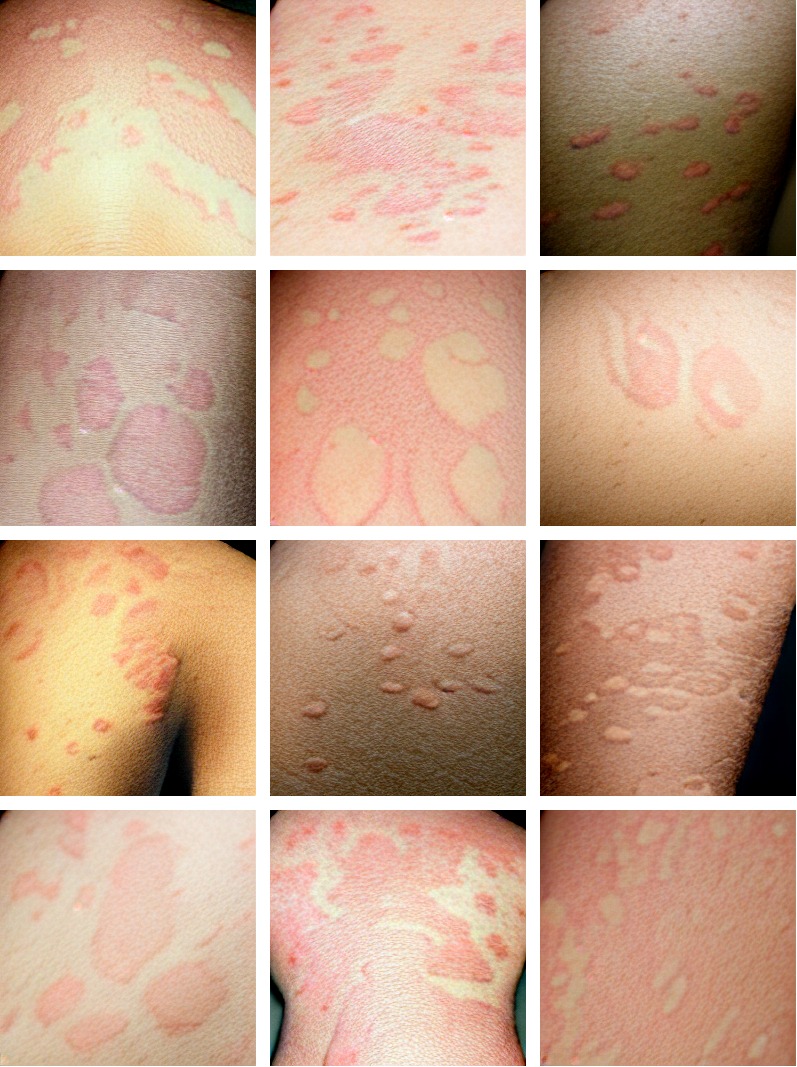}
 \caption{LTI generated samples for \texttt{Urticaria hives} with our structured prompts. }
    \label{fig:gen_supp_2}
\end{figure}

\begin{figure}[t]
    \centering
    \includegraphics[width=14cm]{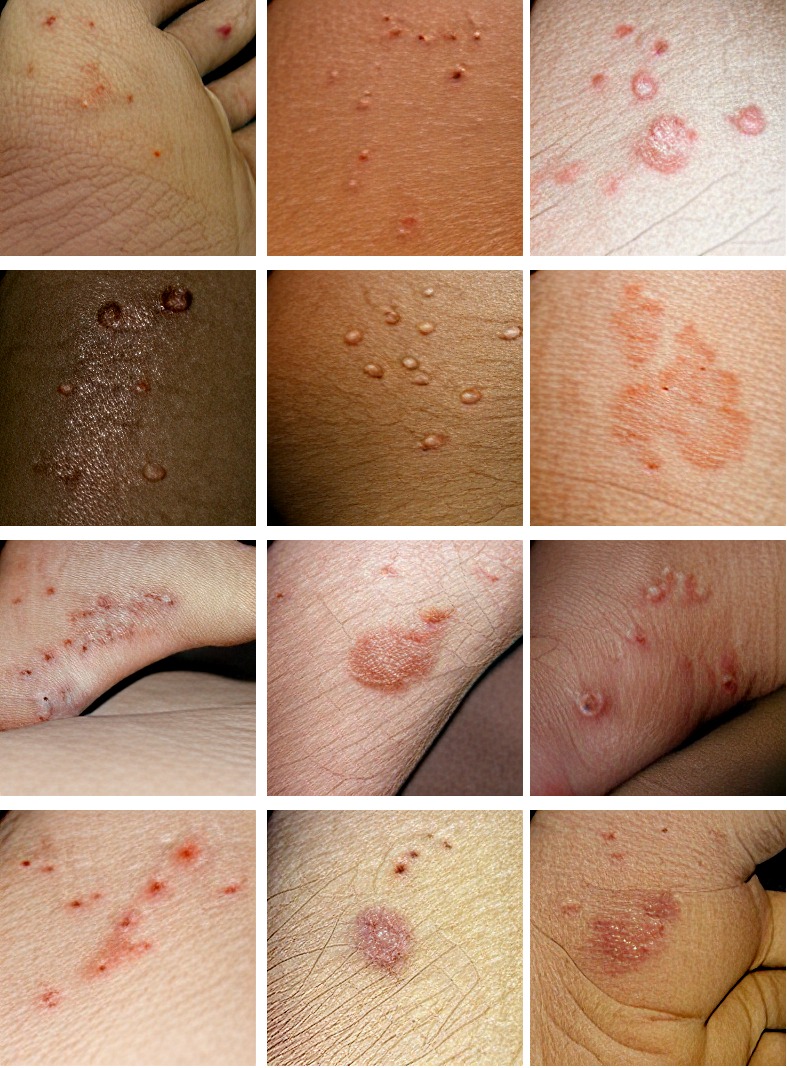}
 \caption{LTI generated samples for \texttt{Scabies} with our structured prompts. }
    \label{fig:gen_supp_3}
\end{figure}

\begin{figure}[t]
    \centering
    \includegraphics[width=14cm]{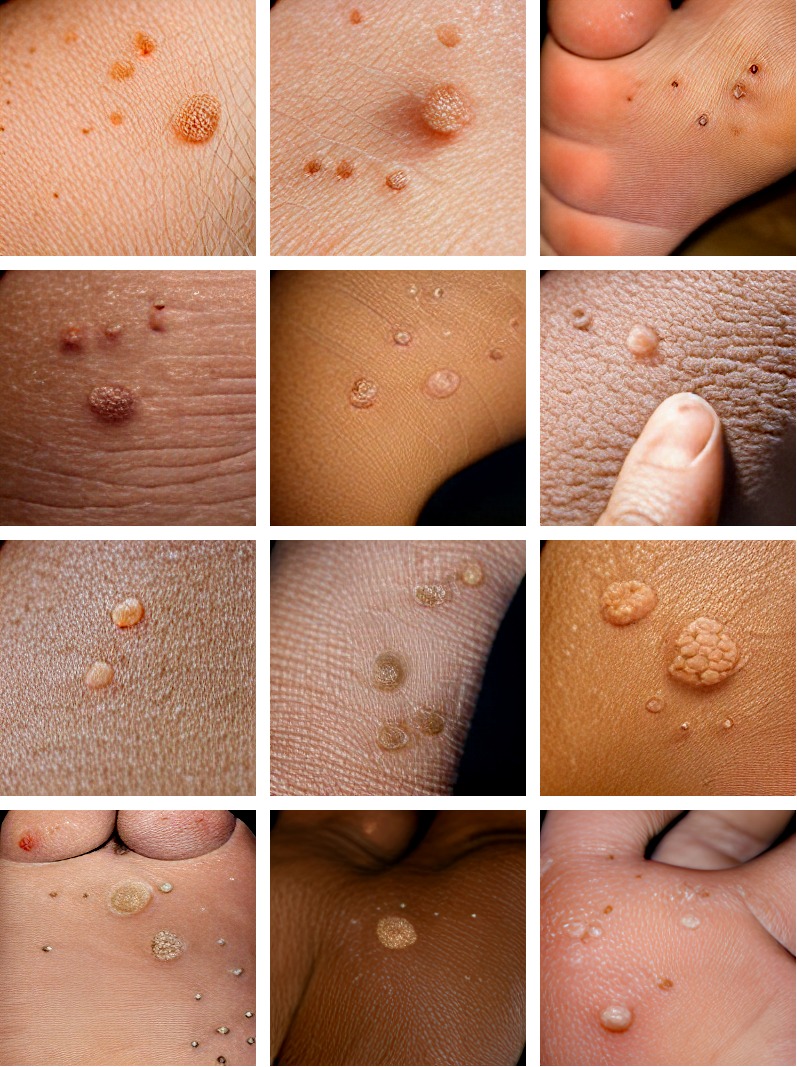}
 \caption{LTI generated samples for \texttt{Warts} with our structured prompts. }
    \label{fig:gen_supp_4}
\end{figure}
\begin{figure}[t]
    \centering
    \includegraphics[width=14cm]{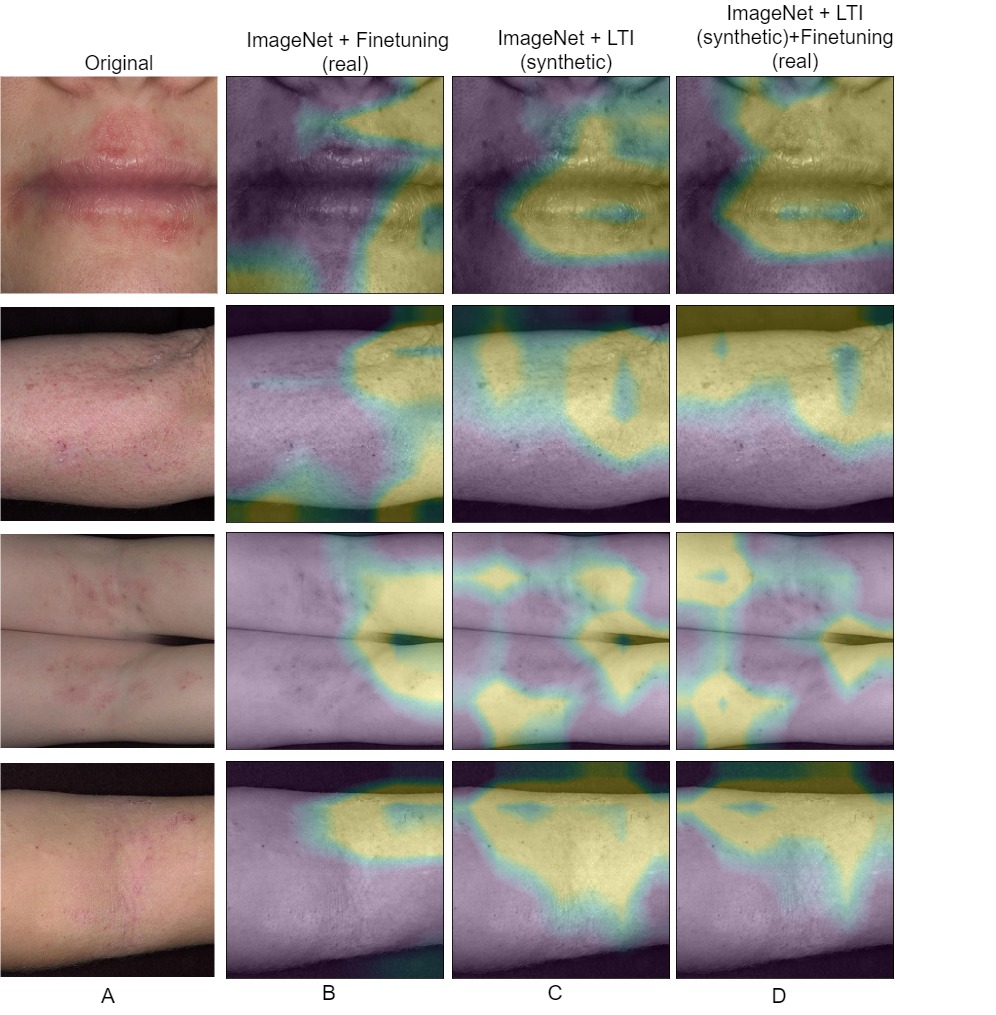}
 \caption{Comparison of the class activation maps for different protocols on {Atopic dermatitis} class -- (A) Images; (B) ImageNet + Finetuning (real); (C) ImageNet + LTI (synthetic); (D) ImageNet + LTI (synthetic) + Finetuning (real). The activation with the synthetic training (C and D) is qualitatively more accurate than the one without it (B).\vspace{-3.0ex} }
    \label{fig:gen_supp_5}
\end{figure}
\begin{figure}[t]
    \centering
    \includegraphics[width=14cm]{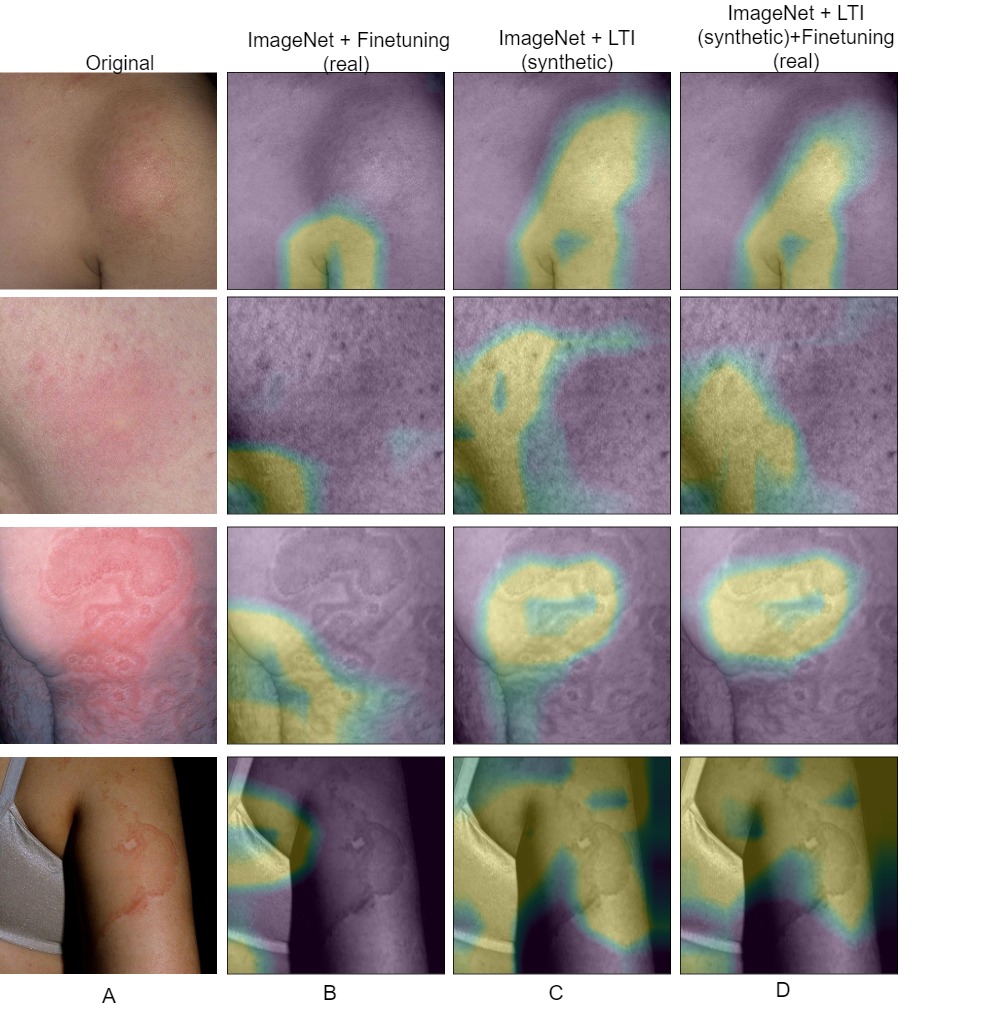}
 \caption{Comparison of the class activation maps for different protocols on \texttt{Urticaria hives} class -- (A) Images; (B) ImageNet + Finetuning (real); (C) ImageNet + LTI (synthetic); (D) ImageNet + LTI (synthetic) + Finetuning (real). The activation with the synthetic training (C and D) is qualitatively more accurate than the one without it (B).\vspace{-3.0ex} }
    \label{fig:gen_supp_6}
\end{figure}
\begin{figure}[t]
    \centering
    \includegraphics[width=14cm]{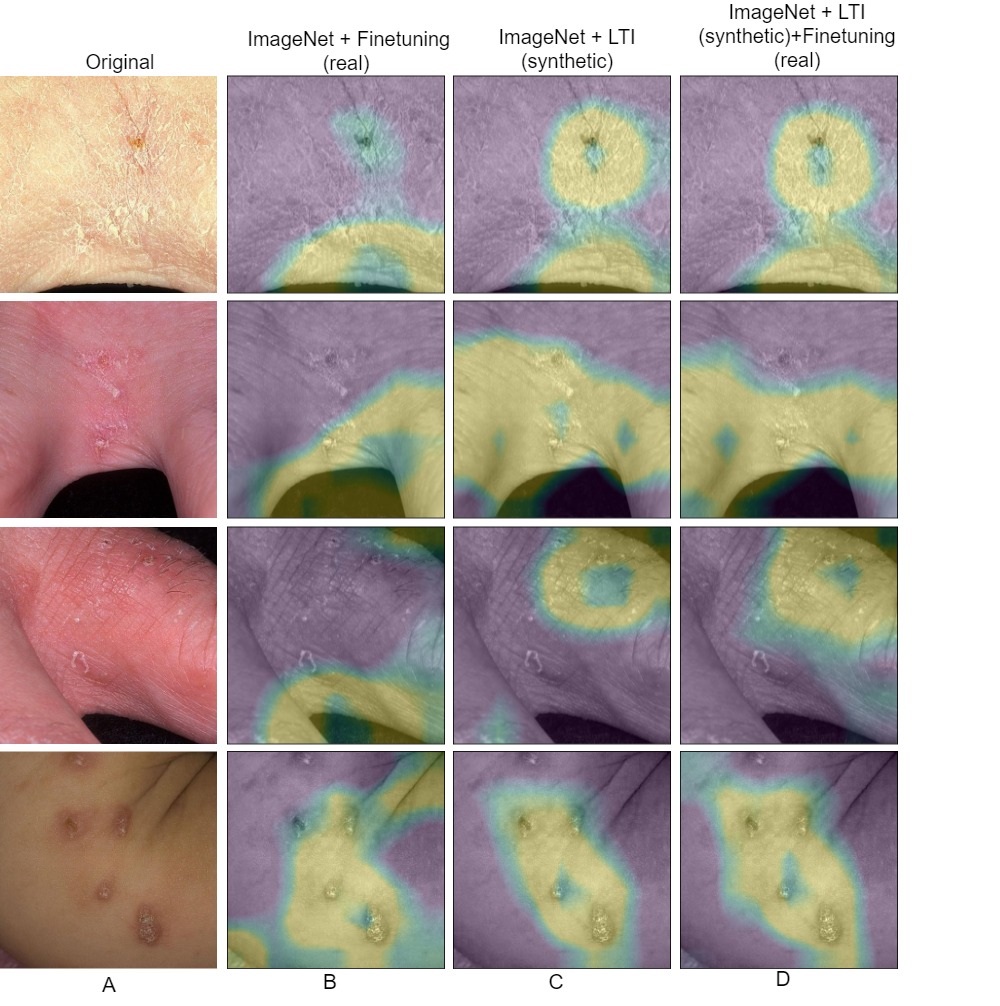}
 \caption{Comparison of the class activation maps for different protocols on \texttt{Scabies} class -- (A) Images; (B) ImageNet + Finetuning (real); (C) ImageNet + LTI (synthetic); (D) ImageNet + LTI (synthetic) + Finetuning (real). The activation with the synthetic training (C and D) is qualitatively more accurate than the one without it (B).\vspace{-3.0ex} }
    \label{fig:gen_supp_7}
\end{figure}
\begin{figure}[t]
    \centering
    \includegraphics[width=14cm]{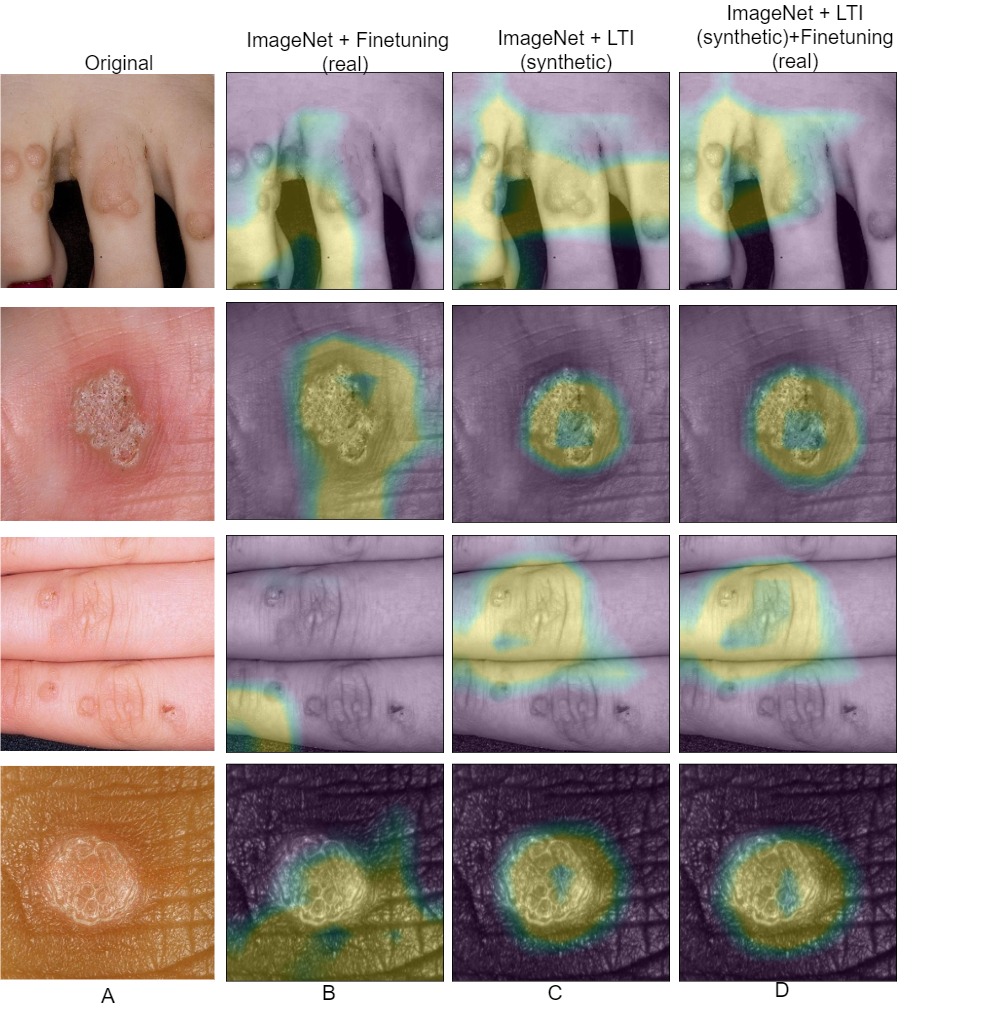}
 \caption{ comparison of the class activation maps for different protocols on  \texttt{Warts} class -- (A) Images; (B) ImageNet + Finetuning (real); (C) ImageNet + LTI (synthetic); (D) ImageNet + LTI (synthetic) + Finetuning (real). The activation with the synthetic training (C and D) is qualitatively more accurate than the one without it (B).\vspace{-3.0ex} }
    \label{fig:gen_supp_8}
\end{figure}

\end{document}